\documentclass[10pt,twocolumn,letterpaper]{article}

\usepackage{wacv}
\usepackage{times}
\usepackage{epsfig}
\usepackage{graphicx}
\usepackage{amsmath}
\usepackage{amssymb}

\usepackage[linesnumbered,ruled,lined,commentsnumbered]{algorithm2e}
\usepackage{caption}
\usepackage{booktabs}
\usepackage{multirow}
\usepackage{tabularx}
\usepackage{subcaption}
\usepackage{multicol}
\usepackage[super]{nth}

\newcommand\blfootnote[1]{%
  \begingroup
  \renewcommand\thefootnote{}\footnote{#1}%
  \addtocounter{footnote}{-1}%
  \endgroup
}

\def\wacvPaperID{781} 

\wacvfinalcopy 

\ifwacvfinal
\def\assignedStartPage{1} 
\fi

\ifwacvfinal
\usepackage[breaklinks=true,bookmarks=false]{hyperref}
\else
\usepackage[pagebackref=true,breaklinks=true,colorlinks,bookmarks=false]{hyperref}
\fi

\ifwacvfinal
\else
\fi

\begin{document}

\title{DynaVSR: Dynamic Adaptive Blind Video Super-Resolution}

\author{\hspace{0cm}Suyoung Lee\textsuperscript{*}\\
\and
\hspace{-3cm}Myungsub Choi\textsuperscript{*}\\
\hspace{-3cm}ASRI, Department of ECE, Seoul National University\\
\hspace{-3cm}{\tt\small \{esw0116, cms6539, kyoungmu\}@snu.ac.kr}\\
\and
\hspace{-3cm}Kyoung Mu Lee\\
}

\maketitle

\begin{abstract}
Most conventional supervised super-resolution (SR) algorithms assume that low-resolution (LR) data is obtained by downscaling high-resolution (HR) data with a fixed known kernel, but such an assumption often does not hold in real scenarios.
Some recent blind SR algorithms have been proposed to estimate different downscaling kernels for each input LR image.
However, they suffer from heavy computational overhead, making them infeasible for direct application to videos.
In this work, we present \textbf{DynaVSR}, a novel meta-learning-based framework for real-world video SR that enables efficient downscaling model estimation and adaptation to the current input.
Specifically, we train a multi-frame downscaling module with various types of synthetic blur kernels, which is seamlessly combined with a video SR network for input-aware adaptation.
Experimental results show that DynaVSR consistently improves the performance of the state-of-the-art video SR models by a large margin, with an order of magnitude faster inference time compared to the existing blind SR approaches.\blfootnote{\hspace{-0.3cm}\textsuperscript{*}indicates equal contribution.}
\end{abstract}

\vspace{-0.5cm}
\section{Introduction}

\vspace{-0.1cm}
Widespread usage of high-resolution (HR) displays in our everyday life has led to increasing popularity of super-resolution (SR) technology, which allows for enhancing the resolution of visual contents from low-resolution (LR) inputs.
Recent advances in deep-learning-based SR approaches for images~\cite{dai2019second,dong2015image,haris2018deep,kim2016accurate,kim2016deeply,lai2017deep,lim2017enhanced,niu2020single,zhang2018residual} and videos~\cite{caballero2017real,jo2018deep,kappeler2016video,tao2017detail,wang2019edvr} are driving this trend, showing excellent performance on public SR benchmarks~\cite{agustsson2017ntire,bevilacqua2012low,huang2015single,martin2001database,nah2019ntire,zeyde2010single}.
However, majority of the models are trained under the assumption that the LR images are downscaled from the ground truth HR images with a \textit{fixed known kernel}, such as MATLAB bicubic.
It has been shown in Shocher \textit{et al.}~\cite{shocher2018zero} that SR performance of the existing models significantly deteriorates if test images do not match such training settings.

SR problem focusing on real-world scenarios with unknown downscaling kernels is called \textit{blind} SR.
Numerous methods have been proposed to accurately estimate image-specific kernels for each input~\cite{bell2019blind,gu2019blind,michaeli2013nonparametric}.
These methods, however, require training the estimation network from scratch at inference time and typically runs in minutes~\cite{bell2019blind,chakrabarti2016neural}, or sometimes up to an hour~\cite{michaeli2013nonparametric} to handle a single image.
Such heavy computational overhead makes the existing approaches impractical to run on a frame-by-frame basis for video SR, as even a short video clip typically contains over hundreds of frames.

\begin{figure*}[t]
    \centering
    \includegraphics[width=1.0\linewidth]{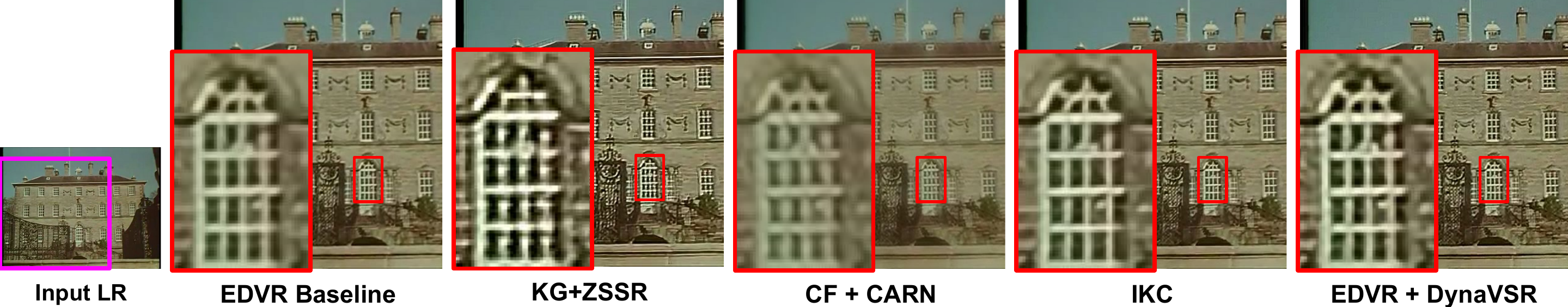}
    \vspace{-0.6cm}
    \caption{Sample results of the proposed DynaVSR on \textit{real} video with unknown degradation kernels.
        The state-of-the-art video SR model (EDVR~\cite{wang2019edvr}) and recent blind SR models (KG (KernelGAN)~\cite{bell2019blind}, CF (CorrectionFilter)~\cite{hussein2020correction}, and IKC~\cite{gu2019blind}) either show blurry outputs or generate unpleasing artifacts, while DynaVSR shows a much clearer result.
    }
    \label{fig:teaser}
    \vspace{-0.5cm}
\end{figure*}

Note that real-world LR video frames contain various different types of degradations, including spatial downsampling and motion blurs.
To solve this problem by learning, we need to collect enough training data for all kinds of degradations, which is computationally infeasible.
However, it can be greatly alleviated if we could effectively estimate the characteristics of the current input video, and build an adaptive model that can adjust its parameters at test time.

In this work, we propose an efficient framework for blind video SR named \textbf{DynaVSR} that can flexibly adapt to dynamic input videos.
Our proposed framework is based on novel downscaling kernel estimation and input-aware adaptation by meta-learning.
It first estimates an approximate downscaling process given input LR video sequences, and generates further downscaled version of the LR frames, which we call \textit{Super LR}, or SLR in short.
Then, using the constructed SLR-LR pairs, the parameters of the video SR (VSR) network as well as the downscaling network are jointly updated.
The final output HR video is obtained by inference through the VSR model with the parameters adapted to the input LR video.
The HR output predicted by DynaVSR for a real-world example sequence is shown in Figure~\ref{fig:teaser}, compared to the recent blind SR methods~\cite{bell2019blind,gu2019blind,hussein2020correction}.
We observe that DynaVSR greatly improves the output quality upon the VSR baselines that are trained with bicubic downsampled data, and shows more visually pleasing results than the existing approaches, even with significantly faster running time (see Sec.~\ref{sec:compute}).

Overall, our contributions are summarized as follows:
\begin{itemize}
    \item We propose DynaVSR, a novel adaptive framework for real-world VSR that combines the estimation of the unknown downscaling process with test-time adaptation via meta-learning.
    \item We greatly reduce the computational complexity of estimating the real downscaling process, thereby enabling real-time execution of SR in videos.
    \item DynaVSR is generic and can be applied to any existing VSR model for consistently improved performance.
\end{itemize}

\section{Related Works}
\vspace{-0.1cm}

While the field of super-resolution (SR) has a long history, in this section, we concentrate on more relevant deep-learning-based approaches and review recent adaptive methods applicable in real-world scenarios.

\subsection{Single-Image SR (SISR)}
Since Dong~\textit{et al.}~\cite{dong2015image} (SRCNN) have shown that a deep learning approach can substantially outperform previous optimization based approaches~\cite{bevilacqua2012low,timofte2013anchored,timofte2014a+,yang2010image,zeyde2010single}, great advances have been made in SISR including VDSR~\cite{kim2016accurate}, EDSR~\cite{lim2017enhanced}, ESRGAN~\cite{ledig2017photo}, and others ~\cite{ahn2018fast,dai2019second,haris2018deep,kim2016deeply,lai2017deep,niu2020single,zhang2018image,zhang2018residual}.
However, despite huge performance boost, many works are limited to only performing well on LR images downscaled by a fixed kernel such as bicubic, and otherwise produce undesirable artifacts.

To overcome this issue, several approaches train SR networks which are applicable to multiple types of degradations, assuming that we already know the degradation kernel (\textit{a.k.a.} non-blind SR).
SRMD~\cite{zhang2018learning} used the LR image and its corresponding degradation kernel as the model inputs to generate high-quality HR images.
ZSSR~\cite{shocher2018zero} instead apply the same kernel used to generate the LR image to make a smaller LR image, then train an image-specific network.
Park \textit{et al.}~\cite{park2020fast} and Soh \textit{et al.}~\cite{soh2020meta} greatly reduce the time required for input-aware adaptation by incorporating meta-learning.
All methods, however, cannot perform well unless we know the exact downscaling kernel, which is unavailable in real-world cases.
To this end, numerous \textit{blind} SR methods have been proposed.

Blind SR methods first estimate the unknown kernels in a self-supervised manner, and then apply the predicted kernels to the non-blind SR models.
Existing kernel estimation approaches either exploit self-similarity~\cite{glasner2009super,huang2015single,zontak2011internal} (with the hypothesis that similar patterns and structures across different scales appear in natural images) or design an iterative self-correction scheme~\cite{gu2019blind,hussein2020correction}.
Michaeli and Irani~\cite{michaeli2013nonparametric} first propose to estimate the downscaling kernel by exploiting the patch-recurrence property of a single image, which is further improved in KernelGAN~\cite{bell2019blind} by utilizing the Internal-GAN~\cite{shocher2018zero}.
IKC~\cite{gu2019blind} introduce an iterative correction scheme and successfully generates high-quality SR images.
Hussein \textit{et al.}~\cite{hussein2020correction} also correct the downscaling kernel for many iterations, and use the final kernel in the same way as their non-blind settings.

Since most of the aforementioned methods require training the model from scratch to estimate an unknown kernel, they suffer from heavy computational overhead at test time.
IKC does not need training at inference, but still requires many iterations for refining its initial output.
On the other hand, our proposed framework directly integrates the input-aware kernel estimation process with video SR models and achieves better results with faster running time, enabling practical application of blind SR techniques to videos.

\begin{figure*}[h!]
    \centering
    \includegraphics[width=0.9\linewidth]{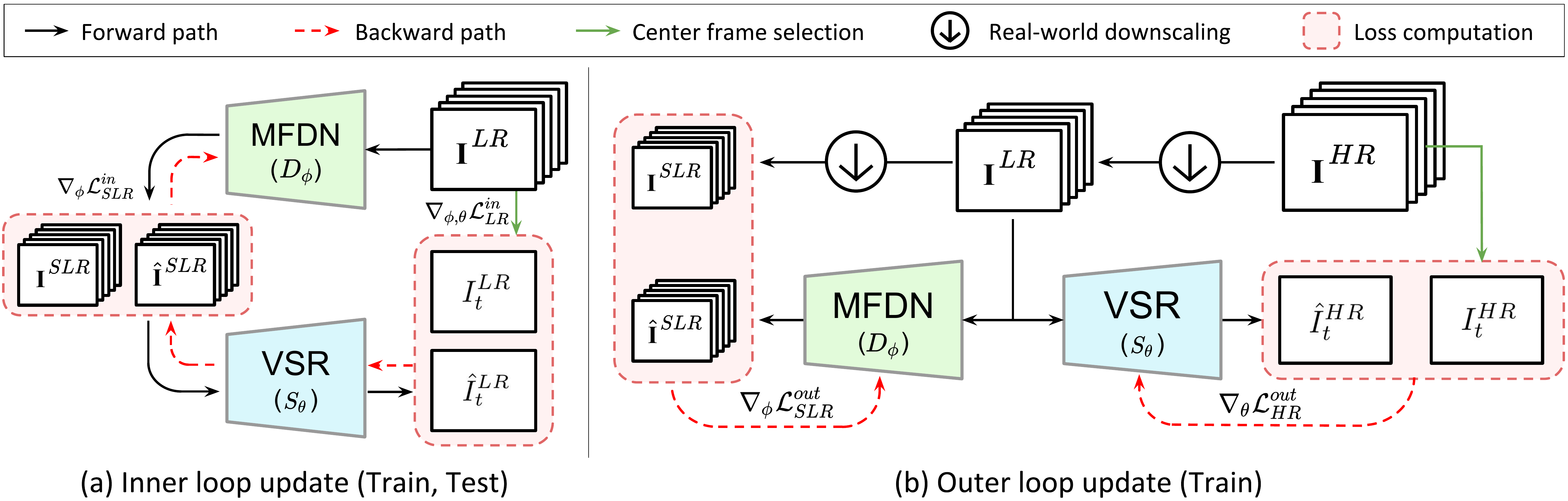}
    \vspace{-0.2cm}
    \caption{Overall training procedure for the proposed DynaVSR framework.
        (a) Both MFDN and VSR network are jointly updated in the inner loop.
        (b) The base parameters, $\phi$ and $\theta$, are separately updated in the outer loop.
    }
    \label{fig:overall}
    \vspace{-0.4cm}
\end{figure*}

\subsection{Video SR (VSR)}
Video SR is different from SISR in that the input frames contain temporal information.
Kappeler \textit{et al.}~\cite{kappeler2016video} first propose a convolutional neural network (CNN) based VSR method by allowing the network input to be a sequence of frames.
Caballero \textit{et al.}~\cite{caballero2017real} and Tao \textit{et al.}~\cite{tao2017detail} incorporate optical flow estimation models to explicitly account for the motion between neighboring frames.
TOFlow~\cite{xue2019video} introduce task-oriented flow, a computationally lighter flow estimation module that is applicable to various video processing tasks.
Since the flow-based methods are highly dependent on the motion estimation accuracy, DUF~\cite{jo2018deep} propose the dynamic upsampling filter network, avoiding explicit calculation of the motion information.
EDVR~\cite{wang2019edvr} also handles motion implicitly with a unified framework, including the Pyramid, Cascading and Deformable convolution (PCD) alignment and the Temporal and Spatial Attention (TSA) fusion processes.
In this work, we use TOFlow, DUF, and EDVR as the baseline VSR models and show how DynaVSR can consistently improve the performance of those models.


\section{Preliminary on MAML}
\label{sec:maml}

Before diving into our main framework, we briefly summarize model-agnostic meta-learning (MAML)~\cite{finn2017model} algorithm that we use for test-time adaptation.
The goal of meta-learning is to rapidly adapt to novel tasks with only few examples.
For MAML, the adaptation process is modeled with a few gradient updates to the parameters.

Specifically, MAML first samples a set of examples $\mathcal{D}_{\mathcal{T}_i}$ from the current task $\mathcal{T}_i \sim p(\mathcal{T})$, where $p(\mathcal{T})$ denotes a distribution of tasks.
Then, adaptation to the current task is done by fine-tuning the model parameters $\theta$:
\begin{equation}
\theta'_i = \theta - \alpha\nabla_{\theta} \mathcal{L}_{\mathcal{T}_i}(S_\theta(\mathcal{D}_{\mathcal{T}_i})),
\label{eq:maml_inner}
\end{equation}
where $\mathcal{L}_{\mathcal{T}_i}$ is a loss function, and $S_\theta$ can be any parameterized model.
After adapting to each task $\mathcal{T}_i$, new examples $\mathcal{D}'_{\mathcal{T}_i}$ are sampled from the same task to test the generalization capability and update the base parameters:
\begin{equation}
\theta  \leftarrow \theta - \beta \nabla_{\theta} \sum_{\mathcal{T}_i}{\mathcal{L}_{\mathcal{T}_i}(S_{\theta'_i}(\mathcal{D}'_{\mathcal{T}_i}))}.
\label{eq:maml_outer}
\end{equation}

Note that Eq.~\ref{eq:maml_inner}, inner loop update, is performed both at training and inference, while the outer loop update (Eq.~\ref{eq:maml_outer}, \textit{a.k.a. meta-update}) is only executed during training.

\section{Dynamic Adaptive Blind VSR Framework}
\label{sec:framework}

In this section, we first summarize the overall framework and the problem formulation, and describe in detail how meta-learning is integrated for efficient adaptation.

\subsection{Overall Framework}

For blind VSR application, we define a \textit{task} as super-resolving each input video sequence.
Since only the input LR frames are available at test time, we further downscale the input to form \textit{super}-low-resolution (SLR) frames.
Then, we can update the model by making it predict the LR frames well given the SLR frames.
The resulting \textit{adapted} parameters perform especially well on the current inputs, generating high-quality HR frames given the LR inputs.

In the \textit{blind} SR setting, each LR input may have come through a different downscaling process, therefore test-time adaptation is crucial.
For video application, real-time execution of estimating the downscaling process is also critical.
Thus, we introduce an efficient Multi-Frame Downscaling Network (MFDN), and combine it with the VSR network.
The proposed framework is named as \textbf{DynaVSR}, as it can adapt well to each of the dynamically-varying input videos.

Figure~\ref{fig:overall} illustrates the overall training process of DynaVSR, which consists of three stages: 1) estimation of the unknown downscaling process with MFDN, 2) joint adaptation of MFDN and VSR network parameters \textit{w.r.t.} each input video, and 3) meta-updating the base parameters for MFDN and VSR network.
At test time, only 1) and 2) are processed, and the updated parameters of the VSR network is used to generate the final super-resolved images.
The detailed training and test processes are described in Sec. ~\ref{sec:meta}.

\subsection{Blind VSR Problem Formulation}
\label{sec:formulation}

The goal of VSR is to accurately predict the HR frames $\left\{\hat{I}^{HR}_t\right\}$ given the input LR frames $\left\{I^{LR}_t\right\}$, where $t$ denotes the time step.
In practice, many recent models such as ~\cite{jo2018deep,tian2020tdan,wang2019edvr} estimate the \textit{center} HR frame $\hat{I}^{HR}_t$ given the surrounding $(2N+1)$ LR frames $I^{LR}_{t\in\mathbb{T}}$, where $\mathbb{T} = \{t-N,\cdots, t+N\}$, and generate the HR sequence in a sliding window fashion.
Thus, VSR problem for a single time step $t$ can be formulated as:
\begin{equation}
    \hat{I}^{HR}_t = S_{\theta}\left(I^{LR}_{t\in\mathbb{T}}\right),
\label{eq:vsr}
\end{equation}

In a fixed-kernel SR setting, a large number of training pairs is available since $\left\{I^{LR}_t\right\}$ can be easily obtained by applying the designated downscaling kernel to $\left\{I^{HR}_t\right\}$.
When tackling \textit{blind} SR, however, the downscaling process is unknown and acquiring a large training set becomes impractical.
As previously studied in KernelGAN~\cite{bell2019blind}, correct estimation of the downscaling process is crucial to the SR performance.
This can be formalized as:
\begin{equation}
    \hat{I}^{LR}_t = D_\phi\left(I^{HR}_t\right),
\label{eq:downscale}
\end{equation}
where $D_\phi$ denotes a downscaling model parameterized by $\phi$.
Existing blind SR approaches typically find a good $D_\phi$ by learning $\phi$ in a self-supervised manner.
Then, using $D_\phi$ that is optimized to the current inputs, the final SR results are obtained in the same way as a non-blind SR setting.

When it comes to blind \textit{video} SR, efficiency becomes the key issue, since videos may contain several hundreds and thousands of frames.
Existing blind \textit{image} SR techniques require long processing time for finding the downscaling model $D_\phi$, and are therefore computationally infeasible (see Sec.~\ref{sec:compute} for runtime comparison).
To this end, we design a new light-weight model, named Multi-Frame Downscaling Network (MFDN), for effective estimation of the downscaling process in real-time.

\subsection{Multi-Frame Downscaling Network (MFDN)}
\label{sec:mfdn}
Though video clips at different time steps can be affected by various different types of degradations (\textit{e.g.} motion blurs, noises), in this work, we primarily focus on the \textit{downscaling} process.
Consequently, we assume that each LR frame $I^{LR}_t$ is generated from the corresponding HR frame $I^{HR}_t$ following the same but unknown downscaling process within a single video sequence.

To model $D_\phi$ in Eq.~\ref{eq:downscale}, we propose MFDN that receives a multi-frame LR inputs and produces the corresponding \textbf{further downscaled} multi-frame outputs.
This process is formulated as:

\begin{equation}
    \hat{I}^{SLR}_{t\in\mathbb{T}} = D_\phi\left(I^{LR}_{t\in\mathbb{T}}\right),
\label{eq:mfdn}
\end{equation}
where $I^{LR}_{t\in\mathbb{T}}$ is an input LR sequence, and $\hat{I}^{SLR}_{t\in\mathbb{T}}$ denotes an estimated \textit{Super LR} (SLR) sequence which is a further downscaled version.
To model various kernels \textit{w.r.t.} different inputs while maintaining efficiency, we model MFDN with a 7-layer CNN including non-linearities.
For handling multi-frame information, 3-D convolutions are used for the first and the last layers of MFDN, and 2-D convolution layers for the rest.
Contrary to the existing methods~\cite{bell2019blind,hussein2020correction}, MFDN does not require additional training at test time.
This greatly improves the efficiency in estimating the unknown downscaling process and thereby enables application to computation-heavy problems like VSR.

To accurately predict the SLR frames for diverse cases with a single discriminative model, MFDN is first pretrained with various synthetic kernels and later employed to our meta-learning framework for further adaptation to each input.
The training LR-SLR pairs are generated by random sampling of the anisotropic Gaussian kernels.
Note that, the LR frames $I^{LR}_{t\in\mathbb{T}}$ are themselves generated from the ground truth HR frames $I^{HR}_{t\in\mathbb{T}}$ by applying randomly selected kernel in the synthetic kernel set.
The corresponding SLR frames $I^{SLR}_{t\in\mathbb{T}}$ are then generated from LR with the same kernel.
Pre-training MFDN is done by minimizing the pixel-wise loss between $I^{SLR}_{t\in\mathbb{T}}$ and the estimated output $\hat{I}^{SLR}_{t\in\mathbb{T}}$.

After pretraining MFDN, it is further fine-tuned during the meta-training process to be readily adaptable to each input.
Though we only use synthetic kernels for training MFDN, it can generalize to the real inputs reasonably well, as we show in the experiments (see Sec.~\ref{sec:qualitative}).
Further analysis on the effects of MFDN is shown in Sec.~\ref{sec:ablation_mfdn}, where we compare it to the other downscaling methods.

\subsection{Meta-Learning for Blind VSR}
\label{sec:meta}

\subsubsection{Meta-training}
For the inner loop update, we first generate $\hat{I}^{SLR}_{t\in\mathbb{T}}$ with MFDN using Eq.~\eqref{eq:mfdn}.
The generated SLR sequence is then fed into the VSR network as the input:
\begin{equation}
    \hat{I}^{LR}_t = S_\theta\left(\hat{I}^{SLR}_{t\in\mathbb{T}}\right) = S_\theta\left(D_\phi\left(I^{LR}_{t\in\mathbb{T}}\right)\right).
\label{eq:inner_forward}
\end{equation}
We introduce two loss terms to update $\phi$ and $\theta$: LR fidelity loss ($\mathcal{L}^{in}_{LR}$) and SLR guidance loss ($\mathcal{L}^{in}_{SLR}$).
The LR fidelity loss ($\mathcal{L}^{in}_{LR}$) indicates the difference between $\hat{I}^{LR}_t$ and $I^{LR}_t$, and we match the type of loss function used for each backbone VSR network (denoted as $L_{VSR}$).
However, $\mathcal{L}^{in}_{LR}$ alone cannot guarantee that the updated MFDN would produce the correct SLR frames.
Inaccurate downscaling estimation can generate erroneous SLR frames, and can also give wrong update signals to the VSR network.
To cope with this issue, SLR guidance loss ($\mathcal{L}^{in}_{SLR}$) is proposed to make sure that MFDN outputs do not move far away from the actual SLR frames.
In practice, $\mathcal{L}^{in}_{SLR}$ is calculated as the $\ell_1$ distance between generated SLR frames $\hat{I}^{SLR}_{t\in\mathbb{T}}$ and the ground truth $I^{SLR}_{t\in\mathbb{T}}$.
The total loss for the inner loop update is computed as a sum of the two terms:
\begin{align}
    \mathcal{L}^{in} &= \mathcal{L}^{in}_{LR} + \mathcal{L}^{in}_{SLR} \\
    &= L_{VSR}\left(\hat{I}^{LR}_t, I^{LR}_t\right) + \left\lVert\hat{I}^{SLR}_{t\in\mathbb{T}} - I^{SLR}_{t\in\mathbb{T}}\right\rVert _1.
\label{eq:inner_loss}
\end{align}
This process corresponds to the left part of Figure~\ref{fig:overall}.

For the outer loop, the \textit{base} parameter values of $\phi$ and $\theta$ (before inner loop updates) are adjusted to make the models more adaptive to new inputs.
Given the input LR sequence $I^{LR}_{t\in\mathbb{T}}$, VSR network and MFDN generate the HR and SLR predictions, correspondingly, as follows:
\begin{equation}
    \hat{I}^{HR}_t = S_{\theta'}\left(I^{LR}_{t\in\mathbb{T}}\right),~~~ \hat{I}^{SLR}_{t\in\mathbb{T}} = D_{\phi'}\left(I^{LR}_{t\in\mathbb{T}}\right).
\label{eq:outer_forward}
\end{equation}
From the two predictions, we can define the two loss terms:
\begin{equation}
   \mathcal{L}^{out}_{HR} = L_{VSR}\left(\hat{I}^{HR}_t, I^{HR}_t\right),
\label{eq:outer_hr_loss}
\end{equation}
\vspace{-0.3cm}
\begin{equation}
   \mathcal{L}^{out}_{SLR} = \left\lVert\hat{I}^{SLR}_{t\in\mathbb{T}} - I^{SLR}_{t\in\mathbb{T}}\right\rVert_1,
\label{eq:outer_slr_loss}
\end{equation}
where each loss is used to update the parameters in corresponding networks.
Note that the loss is calculated with updated parameters, $D_{\phi'}$ and $S_{\theta'}$, but the gradient is calculated \textit{w.r.t.} $\phi$ and $\theta$, respectively.
The right part of Figure~\ref{fig:overall} depicts the outer update mechanism.

\begin{algorithm}[t]
	\SetKwData{Left}{left}\SetKwData{This}{this}\SetKwData{Up}{up}
	\SetKwFunction{Union}{Union}\SetKwFunction{FindCompress}{FindCompress}
	\SetKwInOut{Require}{Require}
	\SetAlgoLined
	
	\Require{$p(\mathcal{T})$: uniform distribution over videos}
	\Require{$\alpha, \beta$: inner / outer-loop learning rates}
	\BlankLine
	Initialize parameters $\theta$ and $\phi$ \\
	\While{not converged}{
	    Sample a batch of sequences $\mathcal{T}_i \sim p(\mathcal{T})$ \\
	    \ForEach{i}{
	        Generate $\left\{I^{HR}_t\right\}_i$, $\left\{I^{LR}_t\right\}_i$, $\left\{I^{SLR}_t\right\}_i$ from $\mathcal{T}_i$ using random synthetic kernels \\
	        Generate $\left\{\hat{I}^{SLR}_t\right\}_i$ using Eq.~\eqref{eq:mfdn} \\
	        Calculate $\nabla_{\phi, \theta} \mathcal{L}^{in}$ using Eq.~\eqref{eq:inner_loss}\\
		    Compute adapted parameters $\phi'$ and $\theta'$ with: \\
		    $\phi' = \phi - \alpha \nabla_{\phi}\mathcal{L}^{in}$,~~~$\theta' = \theta - \alpha \nabla_{\theta}\mathcal{L}^{in}$ \\
		    Save $\left\{I^{HR}_t\right\}_i$, $\left\{I^{SLR}_t\right\}_i$ for meta-update \\
	    }
	    Update $\theta \leftarrow \theta - \beta\nabla_{\theta} \sum_{\mathcal{T}_i} \mathcal{L}^{out}_{HR}$ using Eq.~\eqref{eq:outer_hr_loss}\\
	    Update $\phi \leftarrow \phi - \beta\nabla_{\phi} \sum_{\mathcal{T}_i} \mathcal{L}^{out}_{SLR}$ using Eq.~\eqref{eq:outer_slr_loss}\\
	}
	\caption{DynaVSR training}
	\label{alg:meta_train}
\end{algorithm}

Algorithm~\ref{alg:meta_train} summarizes the full procedure for training DynaVSR.
Compared to the existing blind SISR approaches, the proposed algorithm has multiple advantages:
1) DynaVSR does not require a necessary number of iterations as a hyperparameter, achieving maximum performance with only a single gradient update, leading to improved computational efficiency compared to IKC~\cite{gu2019blind} or CorrectionFilter~\cite{hussein2020correction};
2) DynaVSR is generic and can be applied to any existing VSR models, while the other methods need specific SR network architectures;
3) DynaVSR can handle multiple frames as inputs for video application.

\vspace{-0.2cm}
\subsubsection{Meta-test}
At test time, only the inner loop update is performed to adapt the MFDN and VSR network parameters to the test input frame sequence.
Since there are no ground truth (GT) SLR frames, we replace it with the SLR frames predicted by our pretrained MFDN.
Although we do not use the real GT SLR frames, we show in experiments that the pseudo GT frames generated by MFDN are still valid (see Sec.~\ref{sec:quantitative}).
The final output HR frame $\hat{I}^{HR}_t$ can then be generated using the updated VSR network.

\section{Experiments}

\begin{table*}[t]
\setlength{\tabcolsep}{4pt}
	\centering
	\caption{\textbf{Quantitative results, running time comparison for meta-training with recent VSR models and blind SISR methods}.
	    We evaluate the benefits of DynaVSR algorithm on Vid4~\cite{liu2013bayesian} and REDS-val~\cite{nah2019ntire} dataset.
	    Performance is measured in PSNR (dB).
	    \textcolor{red}{Red} denotes the best performance, and \textcolor{blue}{blue} denotes the second best.
	    The right part indicates the running time to make a single HD frame.
	    DynaVSR shows the shortest time among the existing algorithms on all three VSR baselines.
	    }
	\scalebox{0.9}{
		\begin{tabular}{l l l @{\extracolsep{\fill}\hspace{0.4cm}} c c c @{\hspace{0.5cm}} c c c | c c c}
			\toprule
			& \multicolumn{2}{c}{\multirow[c]{2}[2]{*}{Method}} & \multicolumn{3}{c}{Vid4~\cite{liu2013bayesian}} & \multicolumn{3}{c}{REDS-val~\cite{nah2019ntire}} & \multicolumn{3}{c}{Time (s)} \\
			
			\cmidrule(lr{3\cmidrulekern}){4-6} \cmidrule(lr{\cmidrulekern}){7-9} \cmidrule(lr{\cmidrulekern}){10-12}
			& & & Iso. & Aniso. & Mixed & Iso. & Aniso. & Mixed & Preprocessing & Super-resolution & Total\\
			\midrule
			\midrule 
			
			\parbox[t]{4mm}{\multirow{4}{*}{\rotatebox[origin=c]{90}{Blind SISR~}}} &
			\multicolumn{2}{l}{KG~\cite{bell2019blind} + ZSSR~\cite{shocher2018zero}} & 25.92 & 24.36 & 22.62 & 29.05 & 27.30 & 25.23 & 58.96 & 92.07 & 151.03 \\
			\cmidrule{2-12} 
			
			& \multicolumn{2}{l}{CF~\cite{hussein2020correction} + DBPN~\cite{haris2018deep}\textsuperscript{*}} & 27.30 & 25.61 & 24.03 & 30.37 & 29.30 & 28.02 & 664.77 & 0.03 & 664.78 \\
		    & \multicolumn{2}{l}{CF~\cite{hussein2020correction} + CARN~\cite{ahn2018fast}\textsuperscript{*}} & 27.95 & 26.82 & 25.62 & 30.98 & 30.47 & 29.70 & 106.11 & \textbf{0.02} & 106.13 \\
		    \cmidrule{2-12} 
		    
		    & \multicolumn{2}{l}{IKC~\cite{gu2019blind}} & \textcolor{red}{\textbf{29.46}} & 26.19 & 27.82 & \textcolor{red}{\textbf{34.10}} & 30.11 & 31.41 & - & 2.21 & 2.21 \\

			\midrule
			\midrule

			\parbox[t]{4mm}{\multirow{7}{*}{\rotatebox[origin=c]{90}{Video SR~~~~}}} &
			\multirow{2}{*}{EDVR~\cite{wang2019edvr}}
			& Baseline & 25.35 & 25.84 & 26.27 & 29.14 & 29.66 & 30.21 & - & 0.40 & 0.40 \\
			& & \textbf{DynaVSR}& \textcolor{blue}{\textbf{28.72}} & \textcolor{red}{\textbf{28.81}} &
			\textcolor{red}{\textbf{29.25}} & \textcolor{blue}{\textbf{32.45}} & \textcolor{red}{\textbf{33.41}} & \textcolor{red}{\textbf{33.67}} & 0.88 & 0.40 & \textbf{1.28} \\

			\cmidrule{2-12} 
			& \multirow{2}{*}{DUF~\cite{jo2018deep}}
			& Baseline & 25.26 & 25.70 & 26.47 & 29.01 & 29.38 & 30.14 & - & 1.00 & 1.00 \\
			& & \textbf{DynaVSR}& 27.43 & \textcolor{blue}{\textbf{27.54}} & 27.77 & 31.23 & 31.29 & 31.36 & \textbf{0.30} & 1.00 & 1.30 \\
            \cmidrule{2-12} 
			& \multirow{2}{*}{TOFlow~\cite{xue2019video}}
			& Baseline & 25.27 & 25.66 & 26.48 & 29.04 & 29.46 & 30.40 & - & 0.98 & 0.98 \\
			& & \textbf{DynaVSR}& 27.16 & 27.07 & 27.69 & 31.50 & \textcolor{blue}{\textbf{31.56}} & \textcolor{blue}{\textbf{31.87}} & 0.96 & 0.98 & 1.94 \\
			\cmidrule[0.8pt]{2-12}
			& \multicolumn{2}{l}{Average PSNR gain} & \textbf{+2.48} & \textbf{+2.07} & \textbf{+1.83} & \textbf{+2.66} & \textbf{+2.59} & \textbf{+2.05} & - & - & -\\
            
			\bottomrule
		\end{tabular}
	}
	\label{tb:quantitative}
	\vspace{-0.4cm}
\end{table*}

\subsection{Dataset}

We use three popular VSR datasets for our experiments: REDS~\cite{nah2019ntire}, Vimeo-90K~\cite{xue2019video}, and Vid4~\cite{liu2013bayesian}.
Also, many low-resolution videos are gathered from YouTube to demonstrate the performance of DynaVSR on real-world scenarios.
\textbf{REDS} dataset is composed of 270 videos each consisting of 100 frames, and each frame has 1280 $\times$ 720 spatial resolution.
Out of 270 train-validation videos, we use 266 sequences for training and the other 4 sequences (REDS-val) for testing, following the experimental settings in Wang \textit{et al.}~\cite{wang2019edvr} (EDVR).
Videos from REDS dataset typically contain large and irregular motion, which makes it challenging for VSR.
\textbf{Vimeo-90K} dataset contains 91,707 short video clips, each containing 7 frames.
We use Vimeo-90K only for training, using the training split of 64,612 clips.
Although the resolution of each frame is low (448 $\times$ 256), Vimeo-90K is one of the most frequently used dataset for training VSR models.
\textbf{Vid4} dataset is widely used for evaluation purposes only; many previous works~\cite{haris2019recurrent,tian2020tdan,wang2019edvr} train their models with Vimeo-90K and report their performance on Vid4 dataset, and we follow the same setting.

\subsection{Implementation Details}

DynaVSR can be applied to any deep-learning-based VSR model, and we show its effectiveness using EDVR \cite{wang2019edvr}, DUF~\cite{jo2018deep}, and TOFlow~\cite{xue2019video} as backbone VSR networks.
All models are initialized with pretrained parameters for scale factor $s=2$, with a known downscaling process, MATLAB bicubic downsampling with anti-aliasing.
Separate models are trained for each training dataset, Vimeo-90K and REDS.
We denote these pretrained models as (bicubic) \textbf{\textit{Baseline}}, and report their performance to show how existing approaches using this ideal downscaling kernel fail in synthetic and real-world settings.

When pretraining MFDN and meta-training DynaVSR, diverse kinds of downscaling kernels are used to generate the HR-LR-SLR patches for each iteration.
Specifically, we select $\sigma_1, \sigma_2 \in \mathcal{U}\left[0.2, 2.0\right]$, and $\theta \in \mathcal{U}\left[-\pi, \pi\right]$ independently for randomly generating many anisotropic Gaussian kernels.
More details are described in supplemantary slides.
The source code along with our pretrained models is made public to facilitate reproduction and further research.\footnote[1]{https://github.com/esw0116/DynaVSR}

\subsection{Quantitative Results}
\label{sec:quantitative}

We thoroughly evaluate the performance improvements of DynaVSR \textit{w.r.t.} the bicubic baselines in three different kinds of synthetic blur kernels: isotropic Gaussian, anisotropic Gaussian, and mixed Gaussian.
For all experiments in this section, we report the standard peak signal-to-noise ratio (PSNR).

For \textit{\textbf{isotropic}} Gaussians, we adapt the \textit{Gaussian8} setting from Gu \textit{et al.}~\cite{gu2019blind}, which consists of eight isotropic Gaussian kernels of $\sigma \in \left[0.8, 1.6\right]$ for scale factor 2, originally proposed for evaluating blind image SR methods.
The HR image is first blurred by the Gaussian kernel and then downsampled by bicubic interpolation.
Since it is unclear from Gu \textit{et al.}~\cite{gu2019blind} how to handle the boundary $\sigma$-s, we evaluate on nine different kernel widths including both $\sigma=0.8$ and $\sigma=1.6$ with step size 0.1.
However, isotropic Gaussian kernels are insufficient to represent various types of degradations in the real world.
Thus, we also evaluate DynaVSR on \textbf{\textit{anistropic}} Gaussian settings, where we fix the Gaussian kernel widths to the boundary values of \textit{Gaussian8} so that $\left(\sigma_x, \sigma_y\right) = (0.8, 1.6)$.
Evaluation is done on 4 cases with different rotations ($0^{\circ}, 45^{\circ}, 90^{\circ}$, and $135^{\circ}$), and the average performance is reported.
Lastly, we introduce a \textbf{\textit{mixed}} setting which consists of randomly generated kernels.
Each sequence is individually downscaled by random Gaussian kernels with $\sigma_1, \sigma_2 \in \mathcal{U}\left[0.2, 2.0\right]$ and $\theta \in \mathcal{U}\left[-\pi, \pi\right]$ with direct downsampling. 
Note that the sampled downscaling kernel is kept same for the entire sequence.

Table~\ref{tb:quantitative} compares the results of DynaVSR with its baselines and other blind SR methods.
Compared to the bicubic baseline, DynaVSR consistently improves the performance over all evaluation settings by a large margin (over 2dB on average).
This proves the effectiveness of adaptation via meta-training, since we use the same architecture without introducing any additional parameters.
Compared to blind SISR models, DynaVSR with any baseline VSR network performs favorably against existing methods in general.
IKC performs well in isotropic Gaussian settings, but DynaVSR with EDVR ranks the second while greatly outperforming IKC for the other evaluation settings.
Note that, while IKC is specifically optimized with isotropic Gaussian kernels only, the reported performance for DynaVSR is from our final model trained also with various other kernels including anisotropic and rotated Gaussians.
\blfootnote{\hspace{-0.3cm}\textsuperscript{*}Since CF require too much time for preprocessing, we report the performance for random 10\% of the validation set. For fair comparison, we show the results of the other models on the same 10\% validation set in the supplementary document. We can observe that the overall trend in performance is almost same as the full evaluation.}

\begin{figure*}[t]
 \centering
    \includegraphics[width=1.0\linewidth]{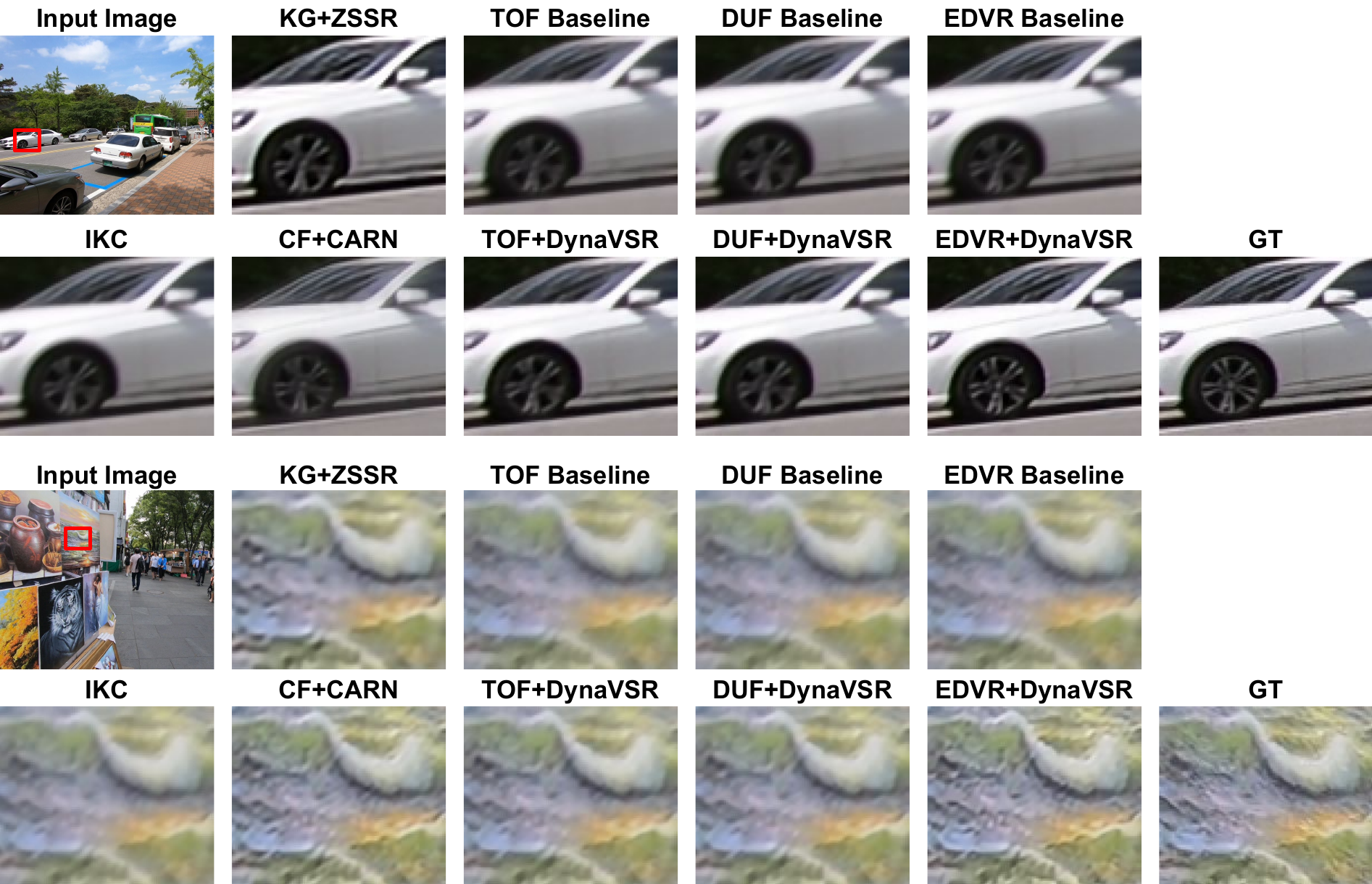}
    \vspace{-0.5cm}
    \caption{\textbf{Qualitative results on REDS-val~\cite{nah2019ntire} dataset.} DynaVSR consistently improves the visual details upon all baseline VSR models, and also produces visually more pleasing outputs compared to recent blind SR approaches.
    }
    \label{fig:qualitative}
    \vspace{-0.4cm}
\end{figure*}

\subsubsection{Time Complexity Analysis}
\label{sec:compute}

The right part of Table~\ref{tb:quantitative} demonstrates the running time for generating a single HD resolution ($1280\times720$) frame from a $\times 2$ downscaled LR frame sequence \textit{w.r.t} each blind SR method.
\textit{Preprocessing} indicates the steps required to prepare the input LR frames for putting through the SR network, which may include kernel estimation (KG~\cite{bell2019blind}) or iterative correction of the inputs to modify their characteristics (CF~\cite{hussein2020correction}).
For IKC~\cite{gu2019blind}, it is difficult to explicitly separate each step, so we include the runtime for iterative correction to \textit{Super-Resolution} category, which reports the inference time for each SR network.

\begin{table}[t]
\setlength{\tabcolsep}{4pt}
	\centering
	\caption{\textbf{Quantitative comparison \textit{w.r.t.} different downscaling models.}
	   We evaluate on REDS-val using EDVR baseline. The right 3 columns indicate the SR performance for each downscaling model in our framework, where the joint training with MFDN shows the best results.
	}
	\scalebox{0.9}{
		\begin{tabular}{l | c | c c c}
			\toprule
             Downscaling models &  SLR & Isotropic  & Anisotropic & Mixed \\
			\midrule
			Bicubic                         & 34.78 & 31.82 & 33.26 & 33.39 \\
			KernelGAN~\cite{bell2019blind}  & 40.84 & 31.99 & 33.31 & 33.48\\
			SFDN          & 45.36 & 32.34 & 33.41 & 33.63 \\
			\textbf{MFDN} & 45.71 & 32.45 & 33.41 & 33.67 \\
			\midrule
			GT                              & $\infty$ & 32.87 & 33.84 & 34.27 \\
			\bottomrule
		\end{tabular}
	}
    \label{tb:ablation_mfdn}
	\vspace{-0.5cm}
\end{table}

Since MFDN is highly efficient, DynaVSR shows much faster preprocessing time compared to existing blind SISR models.
Recent approaches, KG~\cite{bell2019blind} and CF~\cite{hussein2020correction}, require minutes of preprocessing time because both models need to train from scratch at test time, which is very expensive even with a small network.
On the other hand, DynaVSR requires only a single gradient update to the model parameters and successfully reduces the preprocessing time to less than a second (\textbf{more than $\times 60$ faster than KG, and $\times 100$ faster than CF}).
Note that, the preprocessing time for DynaVSR is highly dependent on the architecture of the base VSR network, and the efficiency can be further improved by using more light-weight VSR models.
IKC reports the shortest runtime among the other previous methods, but it still needs multiple iterations for kernel correction, and EDVR+DynaVSR shows more than 40\% shorter runtime.

\subsection{Visual Comparison}
\label{sec:qualitative}

\begin{figure*}[t]
    \centering
    \includegraphics[width=1.0\linewidth]{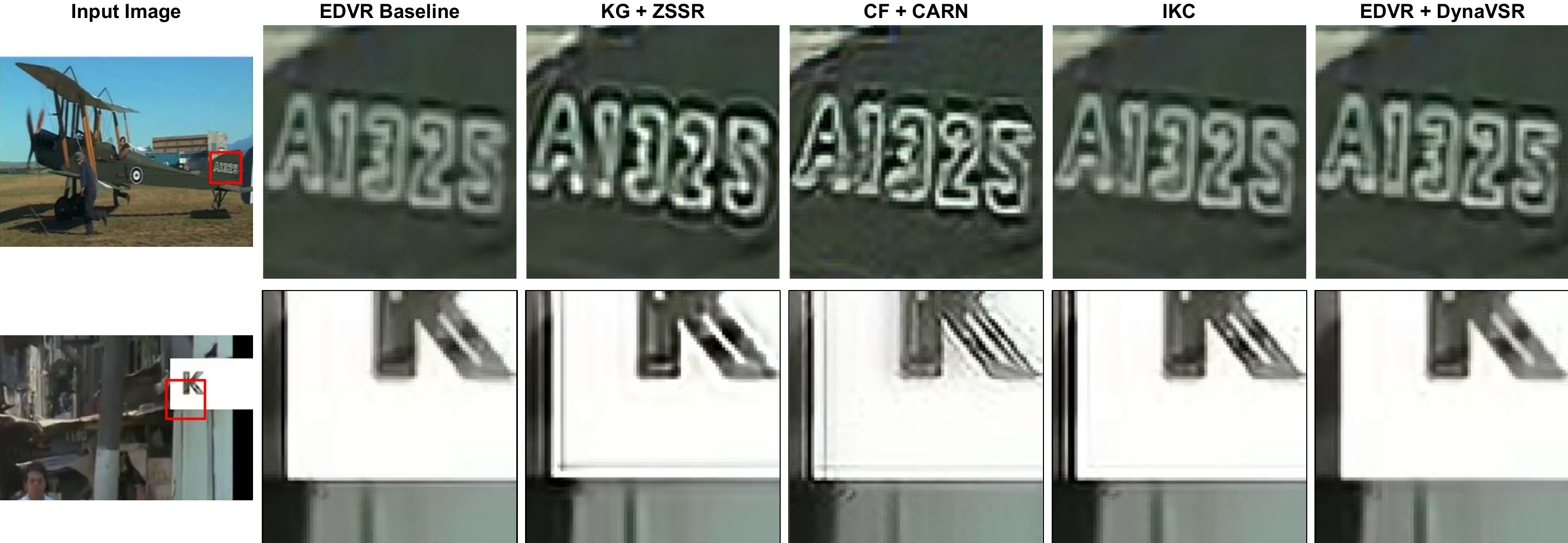}
    \vspace{-0.5cm}
    \caption{\textbf{Qualitative results on \textit{real-world} videos from YouTube} where no ground truth HR frames exist.
        We use EDVR as the base VSR network.
        DynaVSR generates cleaner results while the other blind SISR methods show severe artifacts or blurry outputs.
        Note that, in the second row, all blind SR methods except DynaVSR generate additional boundaries near the corner due to strong ringing artifacts, which does not exist in the original LR frames.
    }
    \label{fig:qualitative_real}
    \vspace{-0.4cm}
\end{figure*}

The qualitative results for REDS-val dataset using random synthetic downscaling kernels are shown in Figure~\ref{fig:qualitative}.
DynaVSR greatly improves the visual quality over all baselines by well adapting to the input frames.
Notably, blurry edges from the bicubic baselines are sharpened, and texture details become much clearer.
Outputs of DynaVSR also show visually more pleasing results compared to three recent blind SISR models, which are shown in the left part of Figure~\ref{fig:qualitative}.
Results for \textit{real-world} low-quality videos from YouTube, where ground truth frames are unavailable, are illustrated in Figure~\ref{fig:teaser},~\ref{fig:qualitative_real}.
Although these videos contain various types of unknown degradations, DynaVSR is robust in producing visually pleasing outputs.
For results on Vid4 dataset, additional results on REDS-val, and more extensive qualitative analysis on many real-world videos, please check our supplementary materials.

\subsection{Analysis}

\subsubsection{Varying Downscaling Models}
\label{sec:ablation_mfdn}

To analyze the effects of end-to-end training of VSR model with a downscaling network, we substitute the MFDN part of DynaVSR with four different downscaling models: bicubic downscaling, KernelGAN~\cite{bell2019blind}, single-frame variant of MFDN (SFDN, Single Frame Downscaling Network), and the ground truth SLR frame downscaled from the LR frame with the correct kernel.
We also evaluate the performance of SLR frame estimation, and report the average PSNR values of VSR network for same settings.
The results are summarized in Table~\ref{tb:ablation_mfdn}.

For generating the SLR frames, we first pretrain MFDN and SFDN until convergence.
SFDN is a single-frame variant of MFDN with the same number of parameters, which regards the temporal dimension as the same as batch dimension, using only 2-D convolutions.
As shown in the leftmost column of Table~\ref{tb:ablation_mfdn}, the MFDN achieves the best performance as a stand-alone downscaler.
We believe that it is due to the multi-frame nature of MFDN, since multiple input frames with similar kernels can be observed to make it easier to recognize the actual downscaling patterns.

The final SR performance after jointly training with the VSR model is also more favorable to MFDN, achieving the closest performance to the ideal case of adapting VSR model with GT SLR frames.

\vspace{-0.1cm}
\subsubsection{Varying the Number of Inner Loop Updates}
We also modify the number of inner loop updates and compare the results.
Table~\ref{tb:ablation_iter} shows the performance while changing the number of iterations within 1, 3, and 5 steps.
The best performance is achieved when we set the number of updates to 1, and more inner loop iterations led to diminishing results on average.
We believe this phenomenon is because of overfitting to the input video sequence and forgetting the general super-resolution capability.
Although the adaptation to each specific input is a crucial step in blind SR problems, it is only beneficial when the model maintains its generalization performance.
Our results show that adaptation with too many inner loop updates drive the model to fall into local optima.
How to regularize the inner loop optimization well to circumvent this issue is beyond the scope of this paper and can be an interesting future direction.

\begin{table}[t]
	\centering
	\caption{\textbf{Effect of the number of inner loop iterations.}
	         We  evaluate  on  REDS-val  using  EDVR.
	         Just 1 inner loop update yields the best performance in general.
	}
	\vspace{-0.2cm}
	\scalebox{0.95}{
		\begin{tabular}{c | c c c}
			\toprule
            \# & Isotropic & Anisotropic & Mixed \\
			\midrule
			1 & 32.45 & \textbf{33.41} & \textbf{33.67} \\
			3 & \textbf{32.75} & 32.38 & 32.96 \\
			5 & 32.17 & 32.23 & 32.57 \\
			\bottomrule
		\end{tabular}
	}
	\label{tb:ablation_iter}
	\vspace{-0.5cm}

\end{table}

\vspace{-0.1cm}
\section{Conclusions}

In this paper, we propose DynaVSR, a novel adaptive blind video SR framework which seamlessly combines the downscaling kernel estimation model into meta-learning-based test-time adaptation scheme in an end-to-end manner.
Compared to existing kernel estimation models for blind SISR, our MFDN extremely improves the computational efficiency and better estimates the downscaling process.
Also, the excessive computation needed for input-aware adaptation of network parameters is minimized to a single gradient update by incorporating meta-learning.
We demonstrate that DynaVSR gives substantial performance gain regardless of the VSR network architecture in various experimental settings including isotropic and anisotropic Gaussian blur kernels.
Furthermore, we empirically show that DynaVSR can be readily applied to real-world videos with unknown downscaling kernels even though it is only trained with synthetic kernels.

{\small
\bibliographystyle{ieee_fullname}
\bibliography{egbib}
}

\clearpage

\setcounter{section}{0}
\setcounter{table}{0}
\setcounter{figure}{0}
\renewcommand{\thesection}{\Alph{section}}
\renewcommand{\thetable}{\Alph{table}}
\renewcommand{\thefigure}{\Alph{figure}}

\onecolumn

\textbf{\begin{center} \Large Supplementary Materials \end{center}}
\vspace{0.3cm}

\section{Implementation Details}

The number of input frames and the loss function in DynaVSR follow the same form as the original VSR method: 5 frames, Charbonnier loss~\cite{lai2017deep} for EDVR, 7 frames with Huber loss for DUF and 7 frames, $\ell_1$ loss for TOFlow.
Since MFDN is fully convolutional, the pre-trained MFDN network can be used regardless of the number of input frames.
We use Adam~\cite{kingma2014adam} optimizer for both the inner and outer loop updates, with the corresponding learning rates of $\alpha=\beta=10^{-5}$.
Training requires 30,000 iterations with a mini-batch size of 4, and $\beta$ is decayed by a factor of 5 at iterations 20,000 and 25,000, while $\alpha$ is kept fixed.
We use full images for Vimeo-90K training, but crop the LR patches of size $128\times128$ for REDS dataset due to GPU memory limits.
Starting from a pretrained VSR baseline, the full meta-training of DynaVSR framework takes approximately 15 hours with a single NVIDIA Geforce RTX 2080 Ti GPU, which includes the 6-hour pretraining step for initializing MFDN parameters.

\section{Additional Quantitative Results}

We provide the SSIM values of Table 3 of the main paper in Table~\ref{tb:full_SSIM}.
Similar to the results for PSNR, EDVR+DynaVSR performs the second best in isotropic Gaussians, and reaches the best in anisotropic and mixed Gaussians.
For all tables in this document, \textcolor{red}{\textbf{red}} denotes the best performance, and \textcolor{blue}{\textbf{blue}} denotes the second best.

Table~\ref{tb:quantitative_iso_vid4},~\ref{tb:quantitative_iso_reds},~\ref{tb:quantitative_aniso_vid4}, and~\ref{tb:quantitative_aniso_reds} provide the detailed quantitative results for all kernel settings that we used to report the average values.
Specifically, Table~\ref{tb:quantitative_iso_vid4} and~\ref{tb:quantitative_iso_reds} show the performance for all 9 $\sigma$-s for the isotropic Gaussian kernels.
Likewise, Table~\ref{tb:quantitative_aniso_vid4} and~\ref{tb:quantitative_aniso_reds} show the performance of all 4 different angles $\theta$ for the anisotropic Gaussian kernels.

Table~\ref{tb:part_psnr} and~\ref{tb:part_ssim} are the results of the 10\% of the validation set, which was used to evaluate Correlation-Filter-based models due to its high computational complexity.
These data provides a fair comparison between the two models using Correction Filter (CF) algorithm, and the other methods in Table 1 of the main paper.
We can observe that the quantitative values are almost same as full evaluation results except for the slight mismatch in the mixed setting, which was originally composed by random sampling of the downscaling kernels that are different for each video sequence.

\subsection{Effect of Inner Loop Learning Rate ($\alpha$)}
The inner loop learning rate $\alpha$ is one of the most import hyperparameters that control the stability of training DynaVSR framework.
If $\alpha$ is too big, the model parameters can jump to a bad local optimum during the inner loop update, resulting in diminishing performance.
On the other hand, if $\alpha$ is too small, the possible performance gain with DynaVSR algorithm may not be fully exploited.
We vary $\alpha$ to $10^{-4}, 10^{-5}$, and $10^{-6}$, and report the effects in Table~\ref{tb:ablation_lr}.
We found that $\alpha=10^{-5}$ gives the best overall results and used this value for all experiments.

\subsection{Larger Scale Factor ($\times 4$ SR)}
Finally, Table~\ref{tb:scale4} shows the $\times 4$ SR results using EDVR.
For larger scaling factor, the size of SLR image will get smaller, and the adaptation using SLR-LR pair becomes more difficult.
However, DynaVSR still makes substantial performance improvements in both dataset.

\begin{table*}[t]
\setlength{\tabcolsep}{3pt}
	\centering
	\caption{\textbf{SSIM results for meta-training with recent VSR models and blind SISR methods}.
	    We evaluate the benefits of DynaVSR algorithm on Vid4~\cite{liu2013bayesian} and REDS-val~\cite{nah2019ntire} dataset.
	    \textcolor{red}{Red} denotes the best performance, and \textcolor{blue}{blue} denotes the second best.}
	\vspace{0.15cm}
	\scalebox{0.95}{
		\begin{tabular}{l l l @{\extracolsep{\fill}\hspace{0.4cm}} c c c @{\hspace{0.5cm}} c c c}
			\toprule
			& \multicolumn{2}{c}{\multirow[c]{2}[2]{*}{Method}} & \multicolumn{3}{c}{Vid4~\cite{liu2013bayesian}} & \multicolumn{3}{c}{REDS-val~\cite{nah2019ntire}} \\
			
			\cmidrule(lr{3\cmidrulekern}){4-6} \cmidrule(lr{\cmidrulekern}){7-9}
			& & & Iso. & Aniso. & Mixed & Iso. & Aniso. & Mixed \\
			\midrule
			\midrule            
			
			\parbox[t]{4mm}{\multirow{4}{*}{\rotatebox[origin=c]{90}{Blind SISR~}}} &
			\multicolumn{2}{l}{KG~\cite{bell2019blind} + ZSSR~\cite{shocher2018zero}} & 0.8395 & 0.8075 & 0.7495 & 0.8673 & 0.8366 & 0.7721 \\
			\cmidrule{2-9} 
			
			& \multicolumn{2}{l}{CF~\cite{hussein2020correction} + DBPN~\cite{haris2018deep}\textsuperscript{*}} & 0.8675 & 0.8353 & 0.7602 & 0.8741 & 0.8696 & 0.8202 \\
		    & \multicolumn{2}{l}{CF~\cite{hussein2020correction} + CARN~\cite{ahn2018fast}\textsuperscript{*}} & 0.8826 & 0.8731 & 0.8388 & 0.8855 & 0.8973 & 0.8788 \\
		    \cmidrule{2-9} 
		    
		    & \multicolumn{2}{l}{IKC~\cite{gu2019blind}} & \textcolor{red}{\textbf{0.9125}} & 0.8308 & 0.8650 & \textcolor{red}{\textbf{0.9361}} & 0.8821 & 0.9030 \\

			\midrule
			\midrule

			\parbox[t]{4mm}{\multirow{7}{*}{\rotatebox[origin=c]{90}{Video SR~~~~}}} &
			\multirow{2}{*}{EDVR~\cite{wang2019edvr}}
			& Baseline & 0.7846 & 0.8227 & 0.8387 & 0.8470 & 0.8729 & 0.8826 \\
			& & \textbf{DynaVSR}& \textcolor{blue}{\textbf{0.9031}} & \textcolor{red}{\textbf{0.8946}} & \textcolor{red}{\textbf{0.9042}} & \textcolor{blue}{\textbf{0.9286}} & \textcolor{red}{\textbf{0.9363}} & \textcolor{red}{\textbf{0.9388}} \\

			\cmidrule{2-9} 
			& \multirow{2}{*}{DUF~\cite{jo2018deep}}
			& Baseline & 0.7815 & 0.8154 & 0.8376 & 0.8426 & 0.8623 & 0.8759 \\
			& & \textbf{DynaVSR}& 0.8600 & \textcolor{blue}{\textbf{0.8756}} & \textcolor{blue}{\textbf{0.8812}} & 0.8935 & \textcolor{blue}{\textbf{0.9023}} & \textcolor{blue}{\textbf{0.9041}} \\
            \cmidrule{2-9} 
			& \multirow{2}{*}{TOFlow~\cite{xue2019video}}
			& Baseline & 0.7815 & 0.8132 & 0.8366 & 0.8431 & 0.8648 & 0.8808 \\
			& & \textbf{DynaVSR}& 0.8509 & 0.8601 & 0.8735 & 0.9031 & \textcolor{blue}{\textbf{0.9087}} & \textcolor{blue}{\textbf{0.9126}} \\
            
			\bottomrule
		\end{tabular}
	}
	
	\hfill\scriptsize{\textsuperscript{*}Since CF takes a long time, we report the performance for random 10\% of the validation set.}

	\label{tb:full_SSIM}
	\vspace{-0.4cm}
\end{table*}

\begin{table*}[t] 
\setlength{\tabcolsep}{3pt}
	\centering
	\caption{\textbf{Quantitative results of DynaVSR combined with recent VSR algorithms in isotropic Gaussian kernels on Vid4}.
	    We evaluate the benefits of DynaVSR algorithm on Vid4~\cite{liu2013bayesian} dataset.}
	\vspace{0.2cm}
	\scalebox{0.75}{
		\begin{tabular}{l | c c c c c c c c c | c}
			\toprule
			\multicolumn{11}{c}{PSNR} \\
			\midrule
    		Method & $\sigma=0.8$ & $\sigma=0.9$ & $\sigma=1.0$ & $\sigma=1.1$ &$\sigma=1.2$ & $\sigma=1.3$ & $\sigma=1.4$ & $\sigma=1.5$ & $\sigma=1.6$ & Average \\
			\midrule
			\midrule
			KG + ZSSR  & 23.70 & 24.39 & 25.56 & 26.21 & 26.67 & 27.00 & 27.03 & 26.65 & 26.06 & 25.92 \\
			CF + DBPN  & 25.62 & 26.92 & 27.94 & 27.81 & 28.00 & 27.86 & 27.65 & 27.20 & 26.68 & 27.30 \\
			CF + CARN  & 27.47 & 28.09 & 28.30 & 28.63 & 28.52 & 28.28 & 27.93 & 27.37 & 26.95 & 27.95 \\
			IKC        & \textcolor{red}{\textbf{30.21}} & \textcolor{red}{\textbf{30.17}} & \textcolor{red}{\textbf{30.06}} & \textcolor{red}{\textbf{29.88}} & \textcolor{red}{\textbf{29.59}} & \textcolor{red}{\textbf{29.26}} & \textcolor{red}{\textbf{28.97}} & \textcolor{red}{\textbf{28.63}} & \textcolor{red}{\textbf{28.37}} & \textcolor{red}{\textbf{29.46}} \\
			\cmidrule(l){1-11}
			EDVR Baseline           & 28.04 & 27.16 & 26.38 & 25.69 & 25.09 & 24.57 & 24.11 & 23.71 & 23.35 & 25.35 \\
		    EDVR + \textbf{DynaVSR} & 29.37 & 29.15 & \textcolor{blue}{\textbf{28.87}} & \textcolor{blue}{\textbf{28.66}} & \textcolor{blue}{\textbf{28.59}} & \textcolor{blue}{\textbf{28.59}} & \textcolor{blue}{\textbf{28.56}} & \textcolor{blue}{\textbf{28.46}} & \textcolor{blue}{\textbf{28.20}} & \textcolor{blue}{\textbf{28.72}} \\
			\cmidrule(l){1-11}
			DUF Baseline           & 27.82 & 26.99 & 26.25 & 25.60 & 25.02 & 24.52 & 24.08 & 23.68 & 23.33 & 25.26 \\
			DUF + \textbf{DynaVSR} & \textcolor{blue}{\textbf{29.48}} & \textcolor{blue}{\textbf{29.16}} & 28.69 & 28.12 & 27.50 & 26.87 & 26.25 & 25.66 & 25.12 & 27.43 \\
			\cmidrule(l){1-11}
			TOF Baseline           & 27.79 & 27.01 & 26.29 & 25.64 & 25.06 & 24.55 & 24.09 & 23.70 & 23.34 & 25.27 \\
			TOF + \textbf{DynaVSR} & 28.90 & 28.59 & 28.20 & 27.75 & 27.26 & 26.73 & 26.20 & 25.65 & 25.12 & 27.16 \\

			\bottomrule
			\toprule
    		\multicolumn{11}{c}{SSIM} \\
			\midrule
    		Method & $\sigma=0.8$ & $\sigma=0.9$ & $\sigma=1.0$ & $\sigma=1.1$ &$\sigma=1.2$ & $\sigma=1.3$ & $\sigma=1.4$ & $\sigma=1.5$ & $\sigma=1.6$ & Average \\
			\midrule
			\midrule
			KG + ZSSR & 0.8083 & 0.8204 & 0.8454 & 0.8529 & 0.8581 & 0.8576 & 0.8528 & 0.8411 & 0.8189 & 0.8395 \\
			CF + DBPN & 0.8195 & 0.8669 & 0.8906 & 0.8901 & 0.8900 & 0.8839 & 0.8724 & 0.8561 & 0.8394 & 0.8675 \\
			CF + CARN & 0.8932 & 0.9002 & 0.9000 & 0.9005 & 0.8954 & 0.8864 & 0.8750 & 0.8552 & 0.8375 & 0.8826 \\
			IKC       & \textcolor{red}{\textbf{0.9206}} & \textcolor{red}{\textbf{0.9203}} & \textcolor{red}{\textbf{0.9196}} & \textcolor{red}{\textbf{0.9181}} & \textcolor{red}{\textbf{0.9153}} & \textcolor{red}{\textbf{0.9116}} & \textcolor{red}{\textbf{0.9077}} & \textcolor{red}{\textbf{0.9024}} & \textcolor{red}{\textbf{0.8965}} & \textcolor{red}{\textbf{0.9125}} \\
			\cmidrule(l){1-11}
			EDVR Baseline           & 0.8839 & 0.8607 & 0.8357 & 0.8096 & 0.7834 & 0.7578 & 0.7332 & 0.7096 & 0.6873 & 0.7846 \\
			EDVR + \textbf{DynaVSR} & 0.9169 & \textcolor{blue}{\textbf{0.9132}} & \textcolor{blue}{\textbf{0.9089}} & \textcolor{blue}{\textbf{0.9057}} & \textcolor{blue}{\textbf{0.9039}} & \textcolor{blue}{\textbf{0.9023}} & \textcolor{blue}{\textbf{0.8993}} & \textcolor{blue}{\textbf{0.8937}} & \textcolor{blue}{\textbf{0.8838}} & \textcolor{blue}{\textbf{0.9031}} \\
			\cmidrule(l){1-11}
			DUF Baseline           & 0.8772 & 0.8551 & 0.8312 & 0.8062 & 0.7809 & 0.7560 & 0.7317 & 0.7085 & 0.6865 & 0.7815 \\
			DUF + \textbf{DynaVSR} & \textcolor{blue}{\textbf{0.9192}} & 0.9114 & 0.9004 & 0.8860 & 0.8686 & 0.8485 & 0.8262 & 0.8024 & 0.7776 & 0.8600 \\
			\cmidrule(l){1-11}
			TOF Baseline           & 0.8757 & 0.8546 & 0.8312 & 0.8065 & 0.7814 & 0.7564 & 0.7321 & 0.7088 & 0.6868 & 0.7815 \\
			TOF + \textbf{DynaVSR} & 0.9053 & 0.8969 & 0.8864 & 0.8736 & 0.8585 & 0.8411 & 0.8213 & 0.7994 & 0.7760 & 0.8509 \\

			\bottomrule
		\end{tabular}
	}
	\label{tb:quantitative_iso_vid4}
	\vspace{-0.3cm}
\end{table*}

\begin{table*}[t] 
\setlength{\tabcolsep}{3pt}
	\centering
	\caption{\textbf{Quantitative results of DynaVSR combined with recent VSR algorithms in isotropic Gaussian kernels on REDS-val}.
	    We evaluate the benefits of DynaVSR algorithm on REDS-val~\cite{nah2019ntire} dataset.}
	\vspace{0.2cm}
	\scalebox{0.75}{
		\begin{tabular}{l | c c c c c c c c c | c}
			\toprule
			\multicolumn{11}{c}{PSNR} \\
			\midrule
    		Method & $\sigma=0.8$ & $\sigma=0.9$ & $\sigma=1.0$ & $\sigma=1.1$ &$\sigma=1.2$ & $\sigma=1.3$ & $\sigma=1.4$ & $\sigma=1.5$ & $\sigma=1.6$ & Average \\
			\midrule
			\midrule
			KG + ZSSR  & 25.62 & 26.81 & 28.12 & 29.31 & 30.29 & 30.64 & 30.58 & 30.30 & 29.80 & 29.05 \\
			CF + DBPN  & 28.97 & 29.74 & 30.05 & 31.01 & 31.18 & 31.19 & 30.85 & 30.48 & 29.85 & 30.37 \\
			CF + CARN  & 31.17 & 31.32 & 31.53 & 31.55 & 31.44 & 31.00 & 30.81 & 30.29 & 29.72 & 30.98 \\
			IKC        & \textcolor{red}{\textbf{34.57}} & \textcolor{red}{\textbf{34.54}} & \textcolor{red}{\textbf{34.50}} & \textcolor{red}{\textbf{34.38}} & \textcolor{red}{\textbf{34.19}} & \textcolor{red}{\textbf{34.02}} & \textcolor{red}{\textbf{33.81}} & \textcolor{red}{\textbf{33.58}} & \textcolor{red}{\textbf{33.29}} & \textcolor{red}{\textbf{34.10}} \\
			\cmidrule(l){1-11}
			EDVR Baseline           & 31.83 & 30.94 & 30.16 & 29.49 & 28.90 & 28.38 & 27.93 & 27.51 & 27.15 & 29.14 \\
		    EDVR + \textbf{DynaVSR} & \textcolor{blue}{\textbf{33.63}} & 32.32 & \textcolor{blue}{\textbf{32.90}} & \textcolor{blue}{\textbf{32.48}} & \textcolor{blue}{\textbf{32.16}} & \textcolor{blue}{\textbf{31.97}} & \textcolor{blue}{\textbf{31.89}} & \textcolor{blue}{\textbf{31.87}} & \textcolor{blue}{\textbf{31.82}} & \textcolor{blue}{\textbf{32.45}} \\
			\cmidrule(l){1-11}
			DUF Baseline           & 31.44 & 30.69 & 30.00 & 29.37 & 28.82 & 28.32 & 27.88 & 27.46 & 27.13 & 29.01 \\
			DUF + \textbf{DynaVSR} & 33.03 & \textcolor{blue}{\textbf{32.79}} & 32.40 & 31.92 & 31.36 & 30.77 & 30.17 & 29.59 & 29.04 & 31.23 \\
			\cmidrule(l){1-11}
			TOF Baseline           & 31.58 & 30.79 & 30.06 & 29.40 & 28.82 & 28.31 & 27.86 & 27.46 & 27.10 & 29.04 \\
			TOF + \textbf{DynaVSR} & 32.61 & 32.52 & 32.35 & 32.09 & 31.73 & 31.31 & 30.82 & 30.30 & 29.76 & 31.50 \\

			\bottomrule
			\toprule
    		\multicolumn{11}{c}{SSIM} \\
			\midrule
    		Method & $\sigma=0.8$ & $\sigma=0.9$ & $\sigma=1.0$ & $\sigma=1.1$ &$\sigma=1.2$ & $\sigma=1.3$ & $\sigma=1.4$ & $\sigma=1.5$ & $\sigma=1.6$ & Average \\
			\midrule
			\midrule
			KG + ZSSR  & 0.8137 & 0.8398 & 0.8623 & 0.8794 & 0.8903 & 0.8922 & 0.8863 & 0.8776 & 0.8643 & 0.8673 \\
			CF + DBPN  & 0.8535 & 0.8723 & 0.8718 & 0.8960 & 0.9020 & 0.8946 & 0.8787 & 0.8611 & 0.8366 & 0.8741 \\
			CF + CARN  & 0.9172 & 0.9157 & 0.9140 & 0.9090 & 0.8985 & 0.8804 & 0.8678 & 0.8454 & 0.8212 & 0.8855 \\
			IKC        & \textcolor{red}{\textbf{0.9446}} & \textcolor{red}{\textbf{0.9438}} & \textcolor{red}{\textbf{0.9429}} & \textcolor{red}{\textbf{0.9409}} & \textcolor{red}{\textbf{0.9379}} & \textcolor{red}{\textbf{0.9350}} & \textcolor{red}{\textbf{0.9314}} & \textcolor{red}{\textbf{0.9271}} & \textcolor{red}{\textbf{0.9217}} & \textcolor{red}{\textbf{0.9361}} \\
			\cmidrule(l){1-11}
			EDVR Baseline           & 0.9155 & 0.8989 & 0.8814 & 0.8635 & 0.8458 & 0.8285 & 0.8120 & 0.7962 & 0.7815 & 0.8470 \\
			EDVR + \textbf{DynaVSR} & \textcolor{red}{\textbf{0.9446}} & \textcolor{blue}{\textbf{0.9411}} & \textcolor{blue}{\textbf{0.9366}} & \textcolor{blue}{\textbf{0.9319}} & \textcolor{blue}{\textbf{0.9278}} & \textcolor{blue}{\textbf{0.9244}} & \textcolor{blue}{\textbf{0.9213}} & \textcolor{blue}{\textbf{0.9176}} & \textcolor{blue}{\textbf{0.9120}} & \textcolor{blue}{\textbf{0.9286}} \\
			\cmidrule(l){1-11}
			DUF Baseline           & 0.9051 & 0.8906 & 0.8750 & 0.8587 & 0.8422 & 0.8259 & 0.8101 & 0.7950 & 0.7806 & 0.8426 \\
			DUF + \textbf{DynaVSR} & \textcolor{blue}{\textbf{0.9327}} & 0.9274 & 0.9201 & 0.9107 & 0.8992 & 0.8859 & 0.8711 & 0.8553 & 0.8389 & 0.8935 \\
			\cmidrule(l){1-11}
			TOF Baseline           & 0.9079 & 0.8926 & 0.8763 & 0.8593 & 0.8424 & 0.8258 & 0.8097 & 0.7944 & 0.7799 & 0.8431 \\
			TOF + \textbf{DynaVSR} & 0.9305 & 0.8275 & 0.9231 & 0.9171 & 0.9093 & 0.8995 & 0.8877 & 0.8743 & 0.8593 & 0.9031 \\

			\bottomrule
		\end{tabular}
	}
	\label{tb:quantitative_iso_reds}
	\vspace{-0.3cm}
\end{table*}

\begin{table*}[t] 
\setlength{\tabcolsep}{3pt}
	\centering
	\caption{\textbf{Quantitative results of DynaVSR combined with recent VSR algorithms in anisotropic Gaussian kernels on Vid4}.
	    We evaluate the benefits of DynaVSR algorithm on Vid4~\cite{liu2013bayesian} dataset.}
	\vspace{0.2cm}
	\scalebox{0.95}{
		\begin{tabular}{l | c c c c | c}
			\toprule
			\multicolumn{6}{c}{PSNR} \\
			\midrule
    		Method & $\theta=0^{\circ}$ & $\theta=45^{\circ}$ & $\theta=90^{\circ}$ & $\theta=135^{\circ}$ & Average\\
			\midrule
			\midrule
			KG + ZSSSR & 25.09 & 23.71 & 25.10 & 23.55 & 24.36 \\
			CF + DBPN  & 26.22 & 25.49 & 25.70 & 25.01 & 25.61 \\
			CF + CARN  & \textcolor{blue}{\textbf{27.54}} & 26.64 & 26.51 & 26.57 & 26.82 \\
			IKC        & 26.10 & 26.12 & 26.50 & 25.97 & 26.17 \\
			\cmidrule(l){1-6}
			EDVR Baseline           & 25.84 & 25.87 & 25.91 & 25.75 & 25.84 \\
		    EDVR + \textbf{DynaVSR} & \textcolor{red}{\textbf{29.65}} & \textcolor{red}{\textbf{28.12}} & \textcolor{red}{\textbf{29.39}} & \textcolor{red}{\textbf{28.06}} & \textcolor{red}{\textbf{28.81}} \\
			\cmidrule(l){1-6}
			DUF Baseline           & 25.68 & 25.71 & 25.81 & 25.60 & 25.70 \\
			DUF + \textbf{DynaVSR} & 27.29 & \textcolor{blue}{\textbf{27.71}} & \textcolor{blue}{\textbf{27.58}} & \textcolor{blue}{\textbf{27.59}} & \textcolor{blue}{\textbf{27.54}} \\
			\cmidrule(l){1-6}
			TOF Baseline           & 25.60 & 25.71 & 25.72 & 25.62 & 25.66 \\
			TOF + \textbf{DynaVSR} & 27.02 & 27.10 & 27.16 & 26.99 & 27.07 \\

			\bottomrule
			\toprule
    		\multicolumn{6}{c}{SSIM} \\
			\midrule
    		Method & $\theta=0^{\circ}$ & $\theta=45^{\circ}$ & $\theta=90^{\circ}$ & $\theta=135^{\circ}$ & Average\\
			\midrule
			\midrule
			KG + ZSSSR & 0.8305 & 0.8257 & 0.8438 & 0.8175 & 0.8075 \\
			CF + DBPN  & 0.8496 & 0.8310 & 0.8401 & 0.8206 & 0.8353 \\
			CF + CARN  & \textcolor{blue}{\textbf{0.8848}} & 0.8699 & 0.8662 & 0.8715 & 0.8731 \\
			IKC        & 0.8338 & 0.8266 & 0.8445 & 0.8181 & 0.8308 \\
			\cmidrule(l){1-6}
			EDVR Baseline           & 0.8320 & 0.8252 & 0.8142 & 0.8195 & 0.8227 \\
			EDVR + \textbf{DynaVSR} & \textcolor{red}{\textbf{0.9179}} & \textcolor{blue}{\textbf{0.8805}} & \textcolor{red}{\textbf{0.9045}} & \textcolor{blue}{\textbf{0.8756}} & \textcolor{red}{\textbf{0.8946}} \\
			\cmidrule(l){1-6}
			DUF Baseline           & 0.8225 & 0.8174 & 0.8102 & 0.8117 & 0.8154 \\
			DUF + \textbf{DynaVSR} & 0.8756 & \textcolor{red}{\textbf{0.8806}} & \textcolor{blue}{\textbf{0.8699}} & \textcolor{red}{\textbf{0.8764}} & \textcolor{blue}{\textbf{0.8756}} \\
 
 			\cmidrule(l){1-6}
			TOF Baseline           & 0.8186 & 0.8158 & 0.8076 & 0.8108 & 0.8132 \\
			TOF + \textbf{DynaVSR} & 0.8640 & 0.8614 & 0.8579 & 0.8569 & 0.8601 \\

			\bottomrule
		\end{tabular}
	}
	\label{tb:quantitative_aniso_vid4}
	\vspace{-0.3cm}
\end{table*}

\begin{table*}[t] 
\setlength{\tabcolsep}{3pt}
	\centering
	\caption{\textbf{Quantitative results of DynaVSR combined with recent VSR algorithms in anisotropic Gaussian kernels on REDS-val}.
	    We evaluate the benefits of DynaVSR algorithm on REDS-val~\cite{nah2019ntire} dataset.}
	\vspace{0.2cm}
	\scalebox{0.95}{
		\begin{tabular}{l | c c c c | c}
			\toprule
			\multicolumn{6}{c}{PSNR} \\
			\midrule
    		Method & $\theta=0^{\circ}$ & $\theta=45^{\circ}$ & $\theta=90^{\circ}$ & $\theta=135^{\circ}$ & Average \\
			\midrule
			\midrule
			KG + ZSSR & 27.65 & 26.66 & 28.59 & 26.28 & 27.30 \\
			CF + DBPN  & 29.51 & 28.66 & 30.16 & 28.88 & 29.30 \\
			CF + CARN  & 30.90 & 29.91 & 30.96 & 30.09 & 30.47 \\
			IKC        & 29.90 & 30.03 & 30.30 & 30.21 & 30.11 \\
			\cmidrule(l){1-6}
			EDVR Baseline           & 29.33 & 29.45 & 30.05 & 29.82 & 29.66 \\
		    EDVR + \textbf{DynaVSR} & \textcolor{red}{\textbf{33.20}} & \textcolor{red}{\textbf{33.10}} & \textcolor{red}{\textbf{33.53}} & \textcolor{red}{\textbf{33.40}} & \textcolor{red}{\textbf{33.41}} \\
			\cmidrule(l){1-6}
			DUF Baseline           & 29.06 & 29.23 & 29.68 & 29.56 & 29.38 \\
			DUF + \textbf{DynaVSR} & 30.84 & 31.30 & 31.33 & 31.69 & 31.29 \\
			\cmidrule(l){1-6}
			TOF Baseline           & 29.15 & 29.28 & 29.76 & 29.63 & 29.46 \\
			TOF + \textbf{DynaVSR} & \textcolor{blue}{\textbf{31.27}} & \textcolor{blue}{\textbf{31.45}} & \textcolor{blue}{\textbf{31.80}} & \textcolor{blue}{\textbf{31.71}} & \textcolor{blue}{\textbf{31.56}} \\

			\bottomrule
			\toprule
    		\multicolumn{6}{c}{SSIM} \\
			\midrule
    		Method & $\theta=0^{\circ}$ & $\theta=45^{\circ}$ & $\theta=90^{\circ}$ & $\theta=135^{\circ}$ & Average \\
			\midrule
			\midrule
			KG + ZSSR   & 0.8465 & 0.8192 & 0.8662 & 0.8143 & 0.8366 \\
			CF + DBPN   & 0.8635 & 0.8587 & 0.8873 & 0.8689 & 0.8696 \\
			CF + CARN   & 0.9007 & 0.8925 & 0.9003 & 0.8958 & 0.8973 \\
			IKC         & 0.8838 & 0.8801 & 0.8841 & 0.8804 & 0.8821 \\
			\cmidrule(l){1-6}
			EDVR Baseline           & 0.8654 & 0.8699 & 0.8797 & 0.8766 & 0.8729 \\
			EDVR + \textbf{DynaVSR} & \textcolor{red}{\textbf{0.9356}} & \textcolor{red}{\textbf{0.9344}} & \textcolor{red}{\textbf{0.9400}} & \textcolor{red}{\textbf{0.9351}} & \textcolor{red}{\textbf{0.9363}} \\
			\cmidrule(l){1-6}
			DUF Baseline           & 0.8549 & 0.8607 & 0.8666 & 0.8669 & 0.8623 \\
			DUF + \textbf{DynaVSR} & 0.8947 & 0.9034 & 0.9026 & 0.9083 & 0.9023 \\
			\cmidrule(l){1-6}
			TOF Baseline           & 0.8582 & 0.8622 & 0.8702 & 0.8685 & 0.8648 \\
			TOF + \textbf{DynaVSR} & \textcolor{blue}{\textbf{0.9045}} & \textcolor{blue}{\textbf{0.9070}} & \textcolor{blue}{\textbf{0.9127}} & \textcolor{blue}{\textbf{0.9105}} & \textcolor{blue}{\textbf{0.9087}} \\

			\bottomrule
		\end{tabular}
	}
	\label{tb:quantitative_aniso_reds}
	\vspace{-0.3cm}
\end{table*}

\begin{table*}[t]
\setlength{\tabcolsep}{3pt}
	\centering
	\caption{\textbf{Quantitative results(PSNR) for meta-training with recent VSR models and blind SISR methods}.
	    We evaluate the benefits of DynaVSR algorithm on Vid4~\cite{liu2013bayesian} and REDS-val~\cite{nah2019ntire} dataset.
	    Only the 10\% of the full data are measured in this table.}
	\vspace{0.15cm}
	\scalebox{0.95}{
		\begin{tabular}{l l l @{\extracolsep{\fill}\hspace{0.4cm}} c c c @{\hspace{0.5cm}} c c c}
			\toprule
			& \multicolumn{2}{c}{\multirow[c]{2}[2]{*}{Method}} & \multicolumn{3}{c}{Vid4~\cite{liu2013bayesian}} & \multicolumn{3}{c}{REDS-val~\cite{nah2019ntire}} \\
			
			\cmidrule(lr{3\cmidrulekern}){4-6} \cmidrule(lr{\cmidrulekern}){7-9}
			& & & Iso. & Aniso. & Mixed & Iso. & Aniso. & Mixed \\
			\midrule
			\midrule            
			
			\parbox[t]{4mm}{\multirow{4}{*}{\rotatebox[origin=c]{90}{Blind SISR~}}} &
			\multicolumn{2}{l}{KG~\cite{bell2019blind} + ZSSR~\cite{shocher2018zero}} & 25.94 & 24.38 & 21.33 & 28.96 & 27.43 & 25.54 \\
			\cmidrule{2-9} 
			
			& \multicolumn{2}{l}{CF~\cite{hussein2020correction} + DBPN~\cite{haris2018deep}} & 27.30 & 25.61 & 24.03 & 30.37 & 29.30 & 28.02 \\
		    & \multicolumn{2}{l}{CF~\cite{hussein2020correction} + CARN~\cite{ahn2018fast}} & 27.95 & 26.82 & 25.62 & 30.98 & 30.47 & 29.70 \\
		    \cmidrule{2-9} 
		    
		    & \multicolumn{2}{l}{IKC~\cite{gu2019blind}} & \textcolor{red}{\textbf{29.49}} & 26.17 & 27.57 & \textcolor{red}{\textbf{34.10}} & 30.11 & 31.42 \\

			\midrule
			\midrule

			\parbox[t]{4mm}{\multirow{7}{*}{\rotatebox[origin=c]{90}{Video SR~~~~}}} &
			\multirow{2}{*}{EDVR~\cite{wang2019edvr}}
			& Baseline & 25.34 & 25.83 & 26.31 & 29.13 & 29.65 & 29.58  \\
			& & \textbf{DynaVSR}& \textcolor{blue}{\textbf{28.70}} & \textcolor{red}{\textbf{28.79}} &
			\textcolor{red}{\textbf{29.41}} & \textcolor{blue}{\textbf{32.43}} & \textcolor{red}{\textbf{33.40}} & \textcolor{red}{\textbf{33.50}} \\

			\cmidrule{2-9} 
			& \multirow{2}{*}{DUF~\cite{jo2018deep}}
			& Baseline & 25.25 & 25.69 & 26.53 & 29.01 & 29.38 & 29.71 \\
			& & \textbf{DynaVSR}& 27.42 & \textcolor{blue}{\textbf{27.53}} & \textcolor{blue}{\textbf{27.97}} & 31.23 & 31.29 & 31.11 \\
            \cmidrule{2-9} 
			& \multirow{2}{*}{TOFlow~\cite{xue2019video}}
			& Baseline & 25.25 & 25.65 & 26.66 & 29.03 & 29.45 & 29.97 \\
			& & \textbf{DynaVSR}& 27.15 & 27.06 & 27.91 & 31.47 & \textcolor{blue}{\textbf{31.52}} & \textcolor{blue}{\textbf{31.69}} \\
            
			\bottomrule
		\end{tabular}
	}
	
	\label{tb:part_psnr}
	\vspace{-0.4cm}
\end{table*}

\begin{table*}[t]
\setlength{\tabcolsep}{3pt}
	\centering
	\caption{\textbf{SSIM results for meta-training with recent VSR models and blind SISR methods}.
	    We evaluate the benefits of DynaVSR algorithm on Vid4~\cite{liu2013bayesian} and REDS-val~\cite{nah2019ntire} dataset.
	    Only the 10\% of the full data are measured in this table.}
	\vspace{0.15cm}
	\scalebox{0.95}{
		\begin{tabular}{l l l @{\extracolsep{\fill}\hspace{0.4cm}} c c c @{\hspace{0.5cm}} c c c}
			\toprule
			& \multicolumn{2}{c}{\multirow[c]{2}[2]{*}{Method}} & \multicolumn{3}{c}{Vid4~\cite{liu2013bayesian}} & \multicolumn{3}{c}{REDS-val~\cite{nah2019ntire}} \\
			
			\cmidrule(lr{3\cmidrulekern}){4-6} \cmidrule(lr{\cmidrulekern}){7-9}
			& & & Iso. & Aniso. & Mixed & Iso. & Aniso. & Mixed \\
			\midrule
			\midrule            
			
			\parbox[t]{4mm}{\multirow{4}{*}{\rotatebox[origin=c]{90}{Blind SISR~}}} &
			\multicolumn{2}{l}{KG~\cite{bell2019blind} + ZSSR~\cite{shocher2018zero}} & 0.8390 & 0.8093 & 0.7131 & 0.8670 & 0.8394 & 0.7704 \\
			\cmidrule{2-9} 
			
			& \multicolumn{2}{l}{CF~\cite{hussein2020correction} + DBPN~\cite{haris2018deep}} & 0.8675 & 0.8353 & 0.7602 & 0.8741 & 0.8696 & 0.8202 \\
		    & \multicolumn{2}{l}{CF~\cite{hussein2020correction} + CARN~\cite{ahn2018fast}} & 0.8826 & 0.8731 & 0.8388 & 0.8855 & 0.8973 & 0.8788 \\
		    \cmidrule{2-9} 
		    
		    & \multicolumn{2}{l}{IKC~\cite{gu2019blind}} & \textcolor{red}{\textbf{0.9125}} & 0.8301 & 0.8554 & \textcolor{red}{\textbf{0.9362}} & 0.8822 & 0.9042 \\

			\midrule
			\midrule

			\parbox[t]{4mm}{\multirow{7}{*}{\rotatebox[origin=c]{90}{Video SR~~~~}}} &
			\multirow{2}{*}{EDVR~\cite{wang2019edvr}}
			& Baseline & 0.7845 & 0.8225 & 0.8481 & 0.8468 & 0.8728 & 0.8700 \\
			& & \textbf{DynaVSR}& \textcolor{blue}{\textbf{0.9026}} & \textcolor{red}{\textbf{0.8943}} & \textcolor{red}{\textbf{0.9059}} & \textcolor{blue}{\textbf{0.9285}} & \textcolor{red}{\textbf{0.9362}} & \textcolor{red}{\textbf{0.9374}} \\

			\cmidrule{2-9} 
			& \multirow{2}{*}{DUF~\cite{jo2018deep}}
			& Baseline & 0.7815 & 0.8150 & 0.8531 & 0.8425 & 0.8624 & 0.8666 \\
			& & \textbf{DynaVSR}& 0.8595 & \textcolor{blue}{\textbf{0.8752}} & \textcolor{blue}{\textbf{0.8906}} & 0.8936 & \textcolor{blue}{\textbf{0.9026}} & \textcolor{blue}{\textbf{0.8999}} \\
            \cmidrule{2-9} 
			& \multirow{2}{*}{TOFlow~\cite{xue2019video}}
			& Baseline & 0.7806 & 0.8124 & 0.8538 & 0.8429 & 0.8645 & 0.8708 \\
			& & \textbf{DynaVSR}& 0.8503 & 0.8594 & 0.8842 & 0.9028 & \textcolor{blue}{\textbf{0.9082}} & \textcolor{blue}{\textbf{0.9095}} \\
            
			\bottomrule
		\end{tabular}
	}
	
	\label{tb:part_ssim}
	\vspace{-0.4cm}
\end{table*}

\begin{table*}[t]
\setlength{\tabcolsep}{3pt}
	\centering
	\caption{Effect of inner loop learning rate $\alpha$ for EDVR on REDS-val.
	}
	\scalebox{1.0}{
		\begin{tabular}{l | c c c}
			\toprule
         $\alpha$  & Isotropic  & Anisotropic & Mixed \\
			\midrule
			$10^{-4}$ & 31.47 & 30.80 & 30.97 \\
			$10^{-5}$ & \textbf{32.45} & \textbf{33.41} & \textbf{33.67} \\
			$10^{-6}$ & 30.96 & 32.94 & 33.26 \\
			\bottomrule
		\end{tabular}
	}
	\label{tb:ablation_lr}
\end{table*}

\begin{table*}[h]
	\centering
	\caption{Additional experimental results for $\times 4$ SR in EDVR~\cite{wang2019edvr}.
	         The best performance is written in \textbf{bold}.
	         Note that DynaVSR shows better performance also in $\times 4$ SR task.}
            \begin{tabular}{l @{\extracolsep{\fill}\hspace{0.4cm}} c c c @{\hspace{0.5cm}} c c c}
			\toprule
			\multirow[c]{2}[2]{*}{Method} & \multicolumn{3}{c}{Vid4~\cite{liu2013bayesian}} & \multicolumn{3}{c}{REDS-val~\cite{nah2019ntire}} \\
			\cmidrule(lr{3\cmidrulekern}){2-4} \cmidrule(lr{\cmidrulekern}){5-7}
			& Iso. & Aniso. & Mixed & Iso. & Aniso. & Mixed \\
			
			\midrule
			\midrule
			EDVR baseline & \textbf{21.14} & 21.53 & 21.11 & 24.80 & 25.22 & 25.29\\
			EDVR+DynaVSR & \textbf{21.14} & \textbf{24.74} & \textbf{23.84} & \textbf{25.81} & \textbf{26.19} & \textbf{27.11}\\
			\bottomrule
		\end{tabular}
	\label{tb:scale4}
	\vspace{-0.7cm}
\end{table*}

\section{Additional Qualitative Results}

For extensive visual comparison both for synthetic and real video examples, please refer to our supplementary slides.
We also demonstrate the effectiveness of the proposed DynaVSR framework with the attached video demo.
Note that, due to the long running time required for KernelGAN~\cite{bell2019blind} and Correction Filter~\cite{hussein2020correction} algorithms, making a full video is too inefficient.
Therefore, DynaVSR is mainly compared with IKC~\cite{gu2019blind} in our video demo.
The video demo can be downloaded by the following URL: ~\url{https://bit.ly/3pisuGj}.

\end{document}